\documentclass[conference]{IEEEtran}
\IEEEoverridecommandlockouts
\usepackage{amsmath,amssymb,amsfonts}
\usepackage{graphicx}
\usepackage{textcomp}
\usepackage[dvipsnames]{xcolor}
\usepackage{booktabs}
\usepackage{pgfplots}
\pgfplotsset{compat=1.18}

\usepackage{algorithm}
\usepackage{algpseudocode}

\usepackage[style=numeric]{biblatex}
\def\BibTeX{{\rm B\kern-.05em{\sc i\kern-.025em b}\kern-.08em
    T\kern-.1667em\lower.7ex\hbox{E}\kern-.125emX}}
\begin{document}

\title{Meta-Learning for Federated Face
Recognition in Imbalanced Data Regimes}

\author{\IEEEauthorblockN{1\textsuperscript{st} Arwin Gansekoele}
\IEEEauthorblockA{
\textit{Centrum Wiskunde \& Informatica}\\
Amsterdam, the Netherlands \\
awg@cwi.nl}
\and
\IEEEauthorblockN{2\textsuperscript{nd} Emiel Hess}
\IEEEauthorblockA{
\textit{Centrum Wiskunde \& Informatica}\\
Amsterdam, the Netherlands \\
ehess77@outlook.com}
\and
\IEEEauthorblockN{3\textsuperscript{rd} Sandjai Bhulai}
\IEEEauthorblockA{\textit{Vrije Universiteit Amsterdam} \\
Amsterdam, the Netherlands \\
s.bhulai@vu.nl}
}
%\author{\IEEEauthorblockN{Anonymous Authors}}

\maketitle

\begin{abstract}
% TODO Improve specificity

The growing privacy concerns surrounding face image data demand new techniques that can guarantee user privacy. One such face recognition technique that claims to achieve better user privacy is Federated Face Recognition (FRR), a subfield of Federated Learning (FL). However, FFR faces challenges due to the heterogeneity of the data, given the large number of classes that need to be handled. To overcome this problem, solutions are sought in the field of personalized FL. This work introduces three new data partitions based on the CelebA dataset, each with a different form of data heterogeneity. It also proposes Hessian-Free Model Agnostic Meta-Learning (HF-MAML) in an FFR setting. We show that HF-MAML scores higher in verification tests than current FFR models on three different CelebA data partitions. In particular, the verification scores improve the most in heterogeneous data partitions.
To balance personalization with the development of an effective global model, an embedding regularization term is introduced for the loss function. This term can be combined with HF-MAML and is shown to increase global model verification performance. Lastly, this work performs a fairness analysis, showing that HF-MAML and its embedding regularization extension can improve fairness by reducing the standard deviation over the client evaluation scores.
\end{abstract}


\section{Introduction}
%\begin{enumerate}
%    \item Introduce the benefits of large data and the challenges in terms of privacy, especially in the context of face recognition. Here, it is a good moment to introduce redacted datasets.
%    \item One potential way to alleviate these issues is the use of Federated Learning techniques such as FedAvg and FedFR, as we can leave data ownership with the users then.
%    \item Major issues arise specifically in face recognition; we have potentially large data heterogeneity and almost no class overlap. Nevertheless, we would like a fair and accurate model.
%    \item To alleviate these issues, we propose the use of federated meta learning. We make the following contributions \begin{enumerate}
%        \item We propose a federated meta-learning algorithm for face recognition with global/local embedding regularization.
%        \item We introduce two new partitions under CelebA that allow us to create a benchmark for personalized learning.
%        \item We empirically demonstrated the value of meta-learning for personalized FFR under heterogenous settings.
%    \end{enumerate}
%\end{enumerate}

% Introductie wat strakker?

During the past decade, significant advances have been made in \emph{artificial intelligence} (AI) and \emph{machine learning} (ML). The surge in available data and computing power has enabled the development of technologies such as large language models, image classification, and image generation. However, these advancements also bring challenges related to data collection and storage. Issues such as data ownership and privacy are becoming more relevant in a world that is becoming increasingly digital. A domain with a particular sensitivity to privacy concerns is \emph{face recognition} (FR). Datasets such as LFW \cite{LFWTech}, MegaFace \cite{nech_level_2017}, MS-Celeb-1M \cite{guo2016msceleb1m}, and IJB-C \cite{maze_iarpa_2018} have been instrumental in advancing FR technology. Of these datasets, both MegaFace and MS-Celeb-1M have been redacted due to privacy issues at the time of writing. Addressing these privacy concerns is essential for the ethical advancement of face recognition technologies.

An approach to using large-scale datasets while respecting privacy constraints is \emph{federated learning} (FL) \cite{mcmahan_communication-efcient_nodate}. By aggregating data from multiple parties, we can share large amounts of information while alleviating privacy issues. In the context of FR, this approach is referred to as \emph{federated face recognition} (FFR). A significant challenge with FFR is that making i.i.d. data assumptions across clients is unrealistic. Different data owners typically have distinct identities, which means that everyone is solving their own local classification problem. Furthermore, identities may vary in terms of attributes depending on the part of the world the data originate from. Ideally, these algorithms should be not only effective but fair as well. Disparities in model performance could discourage participation from certain parties.

This challenge of data heterogeneity is not new in the field of FL \cite{zhao_federated_2018, zhu2021federated}. Attempts to address data heterogeneity can be dubbed personalized federated learning (PFL). Multiple authors have proposed algorithms to address this issue in FL and FFR \cite{aggarwal_fedface_2021, liu_fedfr_2022}. One approach that is underexplored, yet interesting in the context of FFR is meta-learning. Meta-learning algorithms, such as \emph{model-agnostic meta learning} MAML \cite{finn_model-agnostic_2017}, optimize the model in such a way that it can quickly adapt to any new task. MAML has previously been adapted to an FL setting by \cite{fallah_personalized_2020} and \cite{jiang_improving_2019}. Their application in data heterogeneous settings is interesting, as meta-learning provides a natural multi-task framework. In this work, we propose the use of MAML in an FFR setting and analyze its performance. We make the following contributions.
\begin{enumerate}
    \item We introduce the use of meta-learning in the context of FFR and propose an additional regularization term to alleviate the problem of global/local model mismatch;
    \item We introduce two new splits for the dataset CelebA \cite{liu2015faceattributes} that allow us to evaluate FFR models under data heterogeneity;
    \item We evaluate our approach and demonstrate that meta-learning, in the absence of a global dataset, outperforms FedAvg \cite{mcmahan_communication-efcient_nodate} in both TAR@FAR and fairness under data heterogeneity.
\end{enumerate}

\section{Related Work}
% Improve related works 

\paragraph{Federated Face Recognition} While the naming of FFR is new, the CelebA dataset has long been included in FL benchmarks such as LEAF \cite{caldas_leaf_2019}. The introduction of face recognition functions such as CosFace \cite{wang_cosface_2018} in an FL setup followed later \cite{aggarwal_fedface_2021, niu_federated_2021, liu_fedfr_2022}. FedFace \cite{aggarwal_fedface_2021} identified the issue of sensitive information being stored in the classification head of the global model. Another problem occurs when the assumption is made that every client only has one identity to train on; there are no negative samples to balance out the loss. They addressed this by pre-training a classifier on a global dataset and then regularizing the embeddings at the server during FL. FedFR \cite{liu_fedfr_2022} relaxed the assumption of a single identity per client, demonstrating improvements through a multi-task learning approach. However, note that both these approaches assume the availability of a large global dataset to start training. We drop this assumption in our work.

\paragraph{Meta Learning} MAML was first introduced in \cite{finn_model-agnostic_2017} with \cite{fallah_convergence_2020} providing convergence guarantees for this approach. Work by \cite{chen_federated_2019}, \cite{fallah_personalized_2020}, and \cite{jiang_improving_2019} demonstrated that meta-learning can be applied in an FL setting. Work by \cite{jiang_improving_2019} showed that the FedAvg algorithm can be interpreted as a form of meta-learning. Work In work by \cite{yang2023personalized}, the authors proposed an adaptive method to divide clients into groups, where each group is selected to be similar in terms of data heterogeneity. An issue with MAML is that you need to compute an expensive second derivative. That is why \cite{fallah_convergence_2020} proposed Hessian-Free MAML (HF-MAML), which provides an approximation to the second derivative that avoids the need to compute the Hessian. We use this approximation in our meta-learning approach.

%Each client corresponds to a task where the global model serves as a generic model that everyone can use to build their personalized model.

\paragraph{Non-IID Data} The issue of data heterogeneity is well-known in FL. While FedAvg is robust to non-i.i.d. data in certain cases \cite{mcmahan_communication-efcient_nodate}, other papers have reported accuracy reductions of up to 55\% in highly imbalanced data regimes \cite{zhao_federated_2018}. To properly evaluate our models, we need data partitions that allow exploring performance under non-i.i.d. settings. The most common data partition in FFR is one class per client \cite{caldas_leaf_2019}. Another commonly used partition was used by \cite{liu_fedfr_2022, shang_fedfr_2022}, where each client received an equal number of classes $>1$. 

\paragraph{Fairness} An algorithm needs to be fair to be compliant with legislation and for broader adoption. Multiple works have examined how to define fairness in an FL system \cite{li_fair_2020, li_ditto_nodate}. In work by \cite{li_fair_2020}, $q$-FedAvg is proposed to balance the performance across clients. They do so using a $q$ parameter that weights the parameter updates by their loss values. Clients with a lower loss value are less influential for the final parameter update. They also proposed a variant with meta-learning.

\section{Methods}

\subsection{Federated HF-MAML for FFR}
To improve the personalization in an FL setup for FFR, we propose the use of HF-MAML. MAML is model-agnostic, which means it is compatible with any model that uses gradient descent \cite{finn_model-agnostic_2017}. MAML (and HF-MAML) has been shown to work well with classification models \cite{jiang_improving_2019, fallah_personalized_2020, finn_model-agnostic_2017}, but is more complicated in FR since we have to assume that we know the total number of identities of all clients combined beforehand. 

Previous works have shown that one can naturally interpret meta-learning in an FL setup. Meta-learning normally works with different sets of tasks and computes the gradient of the model in such a way that the distance is minimal for all tasks. In FL, one can interchange `task' with `client' and take the same approach. FL is essentially multitask learning where each client forms their own task. To perform meta-learning on gradient-based approaches, one would normally compute the Hessian to minimize the needed gradient update, as this matrix represents the rate of change of functions. However, this is expensive in terms of computation and memory. We make an estimate of the Hessian by sampling multiple datasets and performing the following weight update.
\begin{equation}
    w_{t+1}^k = w^k_t - \beta \nabla_{\tilde{w}_t^k} f(\tilde{w}_t^k, D_k')[I - \alpha \nabla_{w_t^k}^2 f(w_t^k,D_k'')].
\end{equation}
Here, we define $D_k$ as the dataset of client $k$, $w_{t}^k$ the weights at timestep $t$ for client $k$, $f$ the differentiable function. We denote $D_k'$ and $D_k''$ as two extra datasets sampled from client $k$ to get an estimate for the 2nd derivative. We define $\tilde{w}_t^k$ as 
\begin{equation}
    \tilde{w}_t^k = w - \alpha \nabla f(w,D).
\end{equation}
We then approximate $\nabla_{w_t^k}^2 f(w_t^k,D_k'')$ using
\begin{multline}
        \nabla_{w}^2 f(w,D_k'') \nabla_{\tilde{w}} f(\tilde{w},D_k') =  \\ \frac{1}{2\delta}\bigr[\nabla_{w}f(w+\delta \nabla_{\tilde{w}} f(\tilde{w},D_k'),D_k'')\ -\\ \nabla_{w}f(w-\delta \nabla_{\tilde{w}} f(\tilde{w},D_k'),D_k'') \bigr].
\end{multline}

This provides us with the weight update that we need for our approaches. \paragraph{Global Classification} The classification layer of the model refers to the part of the model after the backbone. This module is usually optimized by a loss function, such as Softmax, CosFace, or ArcFace.  If we send all clients these weights as well, they all get access to the embeddings within this module. This poses privacy risks, as it is known that the original face could be reconstructed from the final embedding. As this is often still considered a valid approach, we include this approach in our evaluation suite as a global classification.

Another issue with sharing the global classification layer occurs because not all clients possess data from all classes. Computing, e.g., a softmax over all classes, does not make sense when not all classes are seen during training. That is why if a client knows that they do not possess data from some class $c_i$, they set the logits for the class to $-\infty$. This essentially removes the influence of that class on the final classification problem, meaning they only use the local classes for optimization.

\paragraph{Local Classification}\label{sec:loc_weights}
A safer alternative to maintaining a global classification layer is to keep the classification layer local and not share the weights of this layer with the global model. This approach is intuitive in the context of FFR, as the assumption that no classes are shared between clients is realistic. Instead of sending weight updates for all layers, we only send weight updates for the backbone layer to the global model.

\paragraph{Embedding regularization}\label{sec:emb_reg}
Note that the lack of overlap between classes could cause local models to drift apart. We propose adding an additional loss term to the local classification layer to alleviate the model drift issue. Adding a regularization term to the local loss function in FFR is not new. Both \cite{li_federated_2020} and \cite{liu_fedfr_2022} added regularization terms to reduce the divergence of local weights. Instead of adding a penalty to the divergence between weights, we propose to reduce the divergence between the embeddings generated as in Eq.~\ref{eq:reg_loss}. 
\begin{equation}\label{eq:reg_loss}
    \text{Loss} = f(x) + C (1 - S_c(f_{\text{emb}}^{\text{global}}(x),f_{\text{emb}}^{\text{local}}(x))),
\end{equation}
where $S_c$ refers to the cosine similarity. With the regularization penalty, every client computes the cosine distance between the starting model's embeddings and the current model's embeddings. This term is weighted by a factor of $C$, which allows a trade-off between increasing personalization and reducing model divergence.

\paragraph{Combined algorithm}\label{sec:algorithm}
Algorithm~\ref{alg:final_local} shows the federated HF-MAML algorithm that we propose for FFR. The parts marked dark blue indicate algorithm additions for the implementation of the local classification layer, the parts marked green refer to using a global classification layer, the part marked red represents embedding regularization, and the parts with cyan color are algorithm additions to make HF-MAML work in an FFR setting. It describes our HF-MAML based approach that we use for effective personalized FFR.
\begin{algorithm}
\caption{Federated HF-MAML local regularized}
\label{alg:final_local}
\begin{algorithmic}
    \State \textbf{Server}
        \State \textbf{Initialize} $w_0^{\text{server}}$ on the server
        \If{Local classification layer}
            \State\textcolor{blue}{ \textbf{Initialize} $a_0^k$ as the classification weights on each $\text{client}_k$, $1 \leq k \leq K$}
        \EndIf
        
        \If{Global classification layer}
            \State\textcolor{ForestGreen}{ \textbf{Initialize} $a_0^{\text{server}}$ as the classification weights on the server}
        \EndIf
        \For{$t$ rounds}
            \For{$\text{client}_k$, $1 \leq k \leq K$}
                \State \textbf{Sample} set $D_k$ from $\text{client}_k$ training dataset
                \State $w_t^k = w_t^{\text{server}}$
                \If{Global classification layer}
                    \State\textcolor{ForestGreen}{$a_t^k = a_t^{\text{server}}$}

                \EndIf
                
                \State \textbf{Define} $F(w,a,D) = f(w,a) + \textcolor{red}{C (1 - S_c(f_{\text{emb}}^{\text{global}}(w,D),f_{\text{emb}}^{\text{local}}(w,D)))}$
                \State $(\tilde{w}_{t+1}^k,\textcolor{blue}{\tilde{a}_{t+1}^k}) = (w_t^k,\textcolor{blue}{a_t^k}) - \alpha \nabla F(w_t^k, \textcolor{blue}{a_t^k},D_k)$
                \State \textbf{Sample} set of images $D_k'$ and $D_k''$ from $\text{client}_k$ training dataset different from $D_k$ 
                \State $(w_{t+1}^k,\textcolor{blue}{a_{t+1}^k}) = (w_t^k,\textcolor{blue}{a_t^k}) - \beta \nabla_{\tilde{w}_t^k,\textcolor{blue}{\tilde{a}_t^k}} f(\tilde{w}_t^k,\textcolor{blue}{\tilde{a}_t^k},D_k')[I - \alpha \nabla_{w_t^k,\textcolor{blue}{a_t^k}}^2 f(w_t^k,\textcolor{blue}{a_t^k},D_k'')]$            
                \State \textbf{Send} $w_{t+1}^k$ back to the server
                \If{Global classification layer}
                    \State \textcolor{ForestGreen}{\textbf{Send} $a_{t+1}^k$ back to the server}
                \EndIf
            \EndFor
            \State $w_{t+1} = \sum_{k=1}^K \frac{n_k}{\sum_{i=1}^K n_i} w^k_{t+1}$
                \If{Global classification layer}
                \State \textcolor{ForestGreen}{$a_{t+1} = \sum_{k=1}^K \frac{n_k}{\sum_{i=1}^K n_i} a^k_{t+1}$}
            \EndIf
        \EndFor
    
\end{algorithmic}
\end{algorithm}

\subsection{Data Partitions}
We evaluate three types of data partitions, two of which we propose in this work. We assume the possibility of multiple identities per client, which gives us an equal class partition as a starting point. This partition tries to divide the number of identities and examples as evenly as possible per client. To better explore the effect of data heterogeneity on different algorithms, we propose two new data partitions: the lognormal partition and the attribute-based partition.

\paragraph{Lognormal class partition}\label{sec:log_method}
\begin{table}
\caption{Train and test sizes for each client for the lognormal client partition.}
\label{tab:lognorm_sizes}
\centering
\begin{tabular}{rrr|rrr}
\toprule
client & train data & test data & client & train data & test data  \\ 
\midrule
1      & 134        & 32        & 11     & 264        & 65         \\ 

2      & 1,488       & 376       & 12     & 389        & 99         \\ 

3      & 748        & 184       & 13     & 2,203       & 554        \\ 

4      & 95         & 23        & 14     & 222        & 54         \\ 

5      & 259        & 66        & 15     & 287        & 69         \\ 

6      & 2,585       & 650       & 16     & 276        & 71         \\ 

7      & 27         & 7         & 17     & 4,657       & 1,163       \\ 

8      & 346        & 86        & 18     & 4,307       & 1,076       \\ 

9      & 1,955       & 495       & 19     & 1,326       & 334        \\ 

10     & 220        & 58        & 20     & 773        & 196        \\
\hline
\end{tabular}
\end{table}
\begin{equation}\label{eq:lognormal}
    C \frac{S_i}{\sum_i S_i}, S_i \sim lognormal(\mu,\sigma), \quad 1\leq i \leq n.
\end{equation}
 First, we propose the lognormal class partition as defined in Eq.~\ref{eq:lognormal}. For each client, we draw a sample from the lognormal distribution. We normalize these samples by the sum of all random samples and multiply them by the total number of identities. The result then gives the number of identities a client should be randomly assigned. Using the lognormal distribution in such a way gives us a few clients with many identities and many clients with few identities.
 The idea of using a lognormal distribution has been proposed previously \cite{niu_federated_2021}. However, they instead partition the data itself instead of the identities, which creates a quantity skewness in the data. We found it more intuitive for FFR to create an identity skew, as this classification problem already inherently consists of many classes with few samples per identity.

\paragraph{Attribute-based partition}\label{sec:attr_method}
\begin{table*}
\caption{Different combinations of attributes that are assigned to each client.}
\resizebox{\textwidth}{!}{
\centering
\begin{tabular}{l||llllllllr} 
\toprule
Cl. & Gender & Glasses & Hat & Old/Yng & Hair      & Attr 6         & Attr 7      & Attr 8     & Size  \\ 
\midrule
1   & Male   & Yes     & No  & Young   & All Types &                &             &            & 1,066  \\ 
2   & Male   & Yes     & No  & Old     & Grey      &                &             &            & 1,007  \\ 
3   & Male   & Yes     & No  & Old     & All Types & Chubby         &             &            & 1,001  \\ 
4   & Female & Yes     & No  & Both    & All Types &                &             &            & 925   \\ 
5   & Both   & No      & Yes & Both    & All Types &                &             &            & 1,010  \\ 
6   & Female & No      & No  & Both    & Blond     & Oval Face      & Rosy Cheeks &            & 1,109  \\ 
7   & Male   & No      & No  & Both    & All Types & Goatee         & Not bald    & Smiling    & 1,044  \\ 
8   & Male   & No      & No  & Both    & All Types & No Goatee      & Bald        &            & 1,019  \\ 
9   & Male   & No      & No  & Both    & Gray      & No Goatee      &             &            & 1,040  \\ 
10  & Male   & No      & No  & Old     & All Types & No Goatee      & Bushy Eyeb. &            & 1,062  \\ 
11  & Male   & No      & No  & Old     & No black  & No Goatee      & Bushy Eyeb. &            & 1,007  \\ 
12  & Male   & No      & No  & Old     & No blond  & No Goatee      & Bushy Eyeb. &            & 1,062  \\ 
13  & Male   & No      & No  & Young   & Brown     & No Goatee      & Bushy Eyeb. &            & 1,077  \\ 
14  & Female & No      & No  & Old     & All Types & Oval Face      & Lipstick    & No R. Chks & 1,076  \\ 
15  & Female & No      & No  & Young   & All Types & No Oval Face   & Lipstick    & Rosy Chks  & 1,248  \\ 
16  & Male   & No      & No  & Both    & All Types & No Bushy Eyeb. & Mustache    & No Bushy Eyeb.   & 1,077  \\ 
17  & Male   & No      & No  & Young   & All Types & Beard          & No Mustache & Bushy Eyeb.     & 1,147  \\ 
18  & Male   & No      & No  & Young   & All Types & Beard          & No Mustache &  No Bushy Eyeb.          & 1,108  \\ 
19  & Male   & No      & No  & Young   & Black     & Bags u. Eyes   &             &            & 1,039  \\ 
20  & Female & No      & No  & Old     & All Types & Bags u. Eyes   &             &            & 1,029  \\
\midrule
\end{tabular}
}

\label{tab:attributes}
\end{table*}
Finally, we propose a data partition type based on an attribute partition. The images in the CelebA dataset have not only been labeled based on identity but on more features, such as whether the person has a beard or not, or if the person is wearing make-up. These attributes can be used for classification or to improve face recognition based on the dependence of attributes \cite{hand2016attributes}. Different attributes can be combined to create more specific partitions. For example, a client has a dataset that contains only images of men with beards and glasses.

This data partition is based on the idea of the FEMNIST dataset that is included in the LEAF benchmark. For FEMNIST, each client is assigned handwritten letters according to the writer. This results in each client having data in different handwriting giving a form of feature-based skewness.

% TODO overlap experimental setup toevoegen
% Further explanation of these data partitions is provided in Section~\ref{sec:data}.

\section{Experimental Setup}
\subsection{Data partitioning}\label{sec:data}
This section provides the concrete partitions generated using our approach. We will make these partitions available for future research. In total, there are 20 clients, 15 of which are used for training and 5 for evaluation. A total of 1,500 classes were used for the experiment. Data for each client are divided into a train, validation and test set, in a $70\%/10\%/20\%$ split. We used around $10\%$ of the original dataset due to resource constraints. All samples were cropped using Haar cascades.

\paragraph{Equal class partition}
In the equal class partition, the data set for each client consists of images belonging to 75 randomly assigned identities, without overlap between clients. Since each class has a different sample size, the total number of images per client can differ.

\paragraph{Lognormal class 
partition}

For the lognormal distribution, we used $\mu = 3$ and $\sigma = 3$. We sample a partition once and store this partition for all runs. The amount of test data and training data per client is shown in Tab.~\ref{tab:lognorm_sizes}.

\paragraph{Attribute-based partition}\label{sec:div_attribute}

The CelebA dataset has a set of 40 binary attributes per image. These attributes range from hair color to whether or not the person is wearing glasses. Since some attributes are more common than others, different combinations of attributes are used to create subsets of data that can be assigned to clients. Tab.~\ref{tab:attributes} shows how we have assigned attributes for each client. For example, client 5 has images with both male and female images. It is the only client with people without glasses. They wear hats, they can be both old and young, and the client has images of people with all types of hair color. In short, the focus of client 5 is on people with hats. 
Client 12 only contains images of men who do not wear glasses or hats. The people in the images are old people without blond hair, which means they can have any hair color except blond. The people in the images do not have goatees, which means that all men with goatees are filtered out. Lastly, all have bushy eyebrows. When an attribute is not mentioned it means that the client is agnostic towards this attribute; it has both people with or without this attribute in its dataset. An identity may have images with different attribute distributions. As can be seen in Tab.~\ref{tab:attributes}, dataset sizes are kept as consistent as possible to reduce the effect of quantity skew as much as possible.

\subsection{Model Setup}
The model architecture used to train both the HF-MAML and FL models is shown in Tab.~\ref{tab:norm_arch}. We used the Arcface loss function with $s=8$ and $m=0.5$, which were chosen through tuning. Important to note is that Tab~\ref{tab:norm_arch} depicts the global classification layer scenario with 1,500 identities. For the local classification layer, the output of the Arcface part depends on the number of local identities. The linear layer between ResNet and Arcface is used to cast the ResNet output vector of size 1,000 to the size required for the embedding vector. In this paper, the embedding vector will have a length of 512. This embedding vector is used to compute the cosine similarity. As described previously, this cosine similarity is used for model evaluation and embedding regularization.

\begin{table}[t]

\caption{The model architecture.}
\label{tab:norm_arch}
\centering
\begin{tabular}{lll}
\toprule
Layer type             & In        & Out                             \\ 
\midrule
ResNet 18 (pre-trained) & 3 $\times$ 128 $\times$ 128 & 1,000                            \\
Linear                 & 1,000      & 512                             \\
Arcface ($s=8$, $m=0.5$)    & 512       & 1,500  \\
\bottomrule
\end{tabular}
\end{table}

\begin{table}[t]

\caption{The hyperparameters used for optimization per method.}
\label{tab:hyperparam}
\centering
\begin{tabular}{llll}
\toprule
All              &       & \multicolumn{1}{l}{HF-MAML} &        \\ 
\midrule
Learning rate                              & 0.01  & $\alpha$                                             & 0.01  \\
Momentum                                   & 0.9   & $\beta$                                            & 0.1    \\ 
& & $\delta$ & 0.001\\
\bottomrule
\end{tabular}
\end{table}

For both training and tuning, we use a stochastic gradient descent optimizer with a learning rate of 0.01 and a momentum of 0.9. Tab.~\ref{tab:hyperparam} shows the hyperparameters used for the HF-MAML algorithm. The chosen hyperparameters were found to be sufficient for our experiments.

\subsection{Training and Evaluation}
Training is performed for 30 rounds and local training is performed once on each client for 50 epochs. For the FedAvg algorithm, one epoch means 50 batches of size 64. For the HF-MAML algorithm, we used three sampling rounds with 64 images per sampling round. In each epoch, the algorithm sampled three batches of size 64. Both FL and HF-MAML were tuned to determine the optimal number of epochs used. Using 50 epochs seemed best in both cases. This process was repeated $10$ times with different seeds to obtain our results.

During the evaluation, each training run is evaluated 5 times for the global model and 5 times for the tuned model. 
In scenarios with a local classification layer, the local classification weights are combined with the global backbone weights to tune the model. We report both the average and standard deviation, the latter serving as the backbone for our fairness analysis.

We perform our evaluation in a way similar to \cite{liu_fedfr_2022}. Our evaluations differ from the IARPA Janus Benchmark \cite{maze_iarpa_2018} as we are primarily interested in the performance per client. We perform a stratified split per client, which means that every identity is in both the train and the test set. We sample a batch of pairs where half the pairs have the same identity and the other half are different. The ones of a different origin may originate from other client test sets. Based on these pairs, we measure the verification performance in TAR@FAR.
\section{Results}
\subsection{Convergence analysis}\label{sec:conv_anal}

\begin{figure}
    \centering

        \centering
        \includegraphics[width=0.9\linewidth]{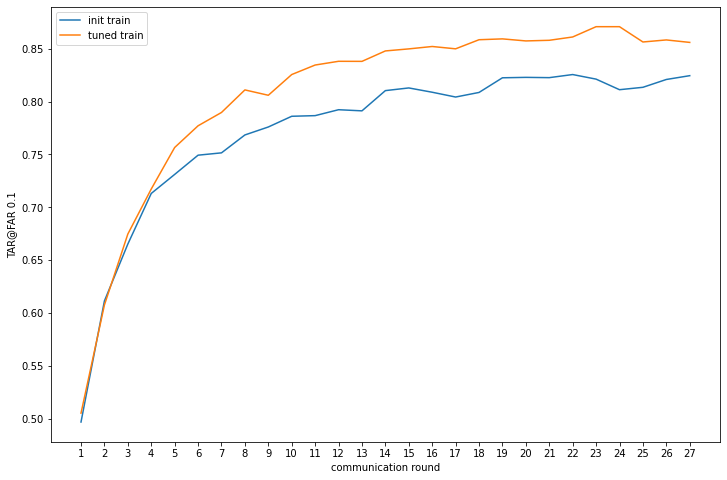} % first figure itself
        \caption{HF-MAML with equal class partition. TAR@FAR 0.1 results after each communication round with the clients 1-15; before and after tuning with 5 batches.}\label{fig:tarfar_hfmaml_uniform_trainclients}
    \end{figure}
    \begin{figure}
        \centering
        \includegraphics[width=0.9\linewidth]{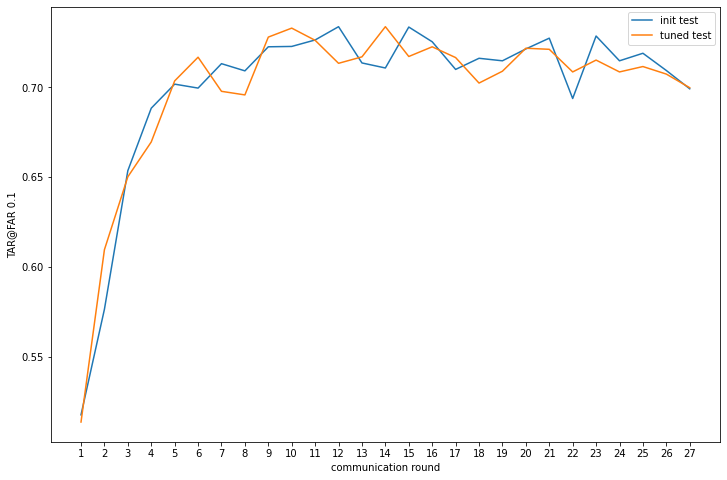} % second figure itself
        \caption{HF-MAML model with equal class partition. TAR@FAR 0.1 results after each communication round with the clients 16-20; before and after tuning with 5 batches.}
    \label{fig:tarfar_hfmaml_uniform_testclients}
\end{figure}

Fig.~\ref{fig:tarfar_hfmaml_uniform_trainclients} shows the average validation score per communication round for the first 15 clients. We compared the global model with the tuned model for both. Figure~\ref{fig:tarfar_hfmaml_uniform_testclients} shows the same validation score per round but for the 5 clients not included during the training. The tuning of the clients included during the training seems to improve the scores, while we do not see this for the clients included later. We found similar training curves for the other methods.

\subsection{Performance Comparison}

%\begin{figure}
%    \centering
    %\includegraphics[width=\linewidth]{sections/results/equal_results.png}
    %\caption{TAR@FAR 0.1 for the equal class partition dataset. Note that FedFR is not tuned and thus the bars are equivalent.}
    %\label{fig:equal_comparison}
%\end{figure}

\begin{table}
    \centering
    \caption{TAR@FAR 0.1 for the equal class partition dataset. Bold implies the best result.}
    \begin{tabular}{lcc}
    \toprule
    & Train & Test \\
    & \multicolumn{2}{c}{Untuned} \\
    \midrule
    FedAvg - Global & $0.783 \pm 0.006$ & $0.710 \pm 0.016$ \\
    FedAvg - Local & $0.779 \pm 0.006$ & $0.718 \pm 0.013$ \\
    HF-MAML - Global & $0.795 \pm 0.007$ & $0.705 \pm 0.016$ \\
    HF-MAML - Local & $\mathbf{0.804} \pm 0.007$ & $\mathbf{0.737} \pm 0.014$ \\ \toprule
    & \multicolumn{2}{c}{Tuned} \\
    \midrule
    FedAvg - Global & $0.836 \pm 0.006$ & $0.760 \pm 0.012$ \\
    FedAvg - Local & $0.817 \pm 0.006$ & $0.740 \pm 0.013$ \\
    HF-MAML - Global & $\mathbf{0.855} \pm 0.005$ & $\mathbf{0.762} \pm 0.013$ \\
    HF-MAML - Local & $0.852 \pm 0.006$ & $0.768 \pm 0.013$ \\ \bottomrule
    \end{tabular}
    \label{tab:equal-comparison}
\end{table}

First, we compared the global and local models for both FedAvg and HF-MAML in Tab.~\ref{tab:equal-comparison} under an equal class partition. We observed that HF-MAML achieves a better TAR@FAR than FedAvg, especially after tuning locally. We also saw little difference between the local and global models, which implies that it is sufficient to keep the classification layer local.

%\begin{figure}
%    \centering
%    \includegraphics[width=\linewidth]{sections/results//lognorm_results.png}
%    \caption{Results for the lognormal class partition dataset. Note that FedFR is not tuned, and thus, the bars are equivalent.}
%    \label{fig:lognorm_comparison}
%\end{figure}

\begin{table}
    \centering
    \caption{TAR@FAR0.1 for the lognormal class partition dataset. Bold implies the best result.}
    \begin{tabular}{lcc}
    \toprule
    & Train & Test \\
    & \multicolumn{2}{c}{Untuned} \\
    \midrule
    FedAvg - Global & $0.707 \pm 0.013$ & $0.675 \pm 0.013$ \\
    FedAvg - Local & $0.715 \pm 0.013$ & $0.686 \pm 0.013$ \\
    FedAvg - Regularization & $0.709 \pm 0.009$ & $0.704 \pm 0.013$ \\
    HF-MAML - Global & $0.722 \pm 0.012$ & $0.695 \pm 0.012$ \\
    HF-MAML - Local & $0.725 \pm 0.014$ & $0.705 \pm 0.016$ \\ 
    HF-MAML - Regularization & $\mathbf{0.734} \pm 0.011$ & $\mathbf{0.726} \pm 0.011$ \\ \toprule
    & \multicolumn{2}{c}{Tuned} \\
    \midrule
    FedAvg - Global & $0.788 \pm 0.012$ & $0.715 \pm 0.016$ \\
    FedAvg - Local & $0.786 \pm 0.013$ & $0.707 \pm 0.014$ \\
    FedAvg - Regularization & $0.773 \pm 0.008$ & $0.715 \pm 0.017$ \\
    HF-MAML - Global & $\mathbf{0.810} \pm 0.014$ & $\mathbf{0.737} \pm 0.013$ \\
    HF-MAML - Local & $0.802 \pm 0.014$ & $0.731 \pm 0.012$ \\ 
    HF-MAML - Regularization & $0.798 \pm 0.011$ & $0.735 \pm 0.013$ \\ \bottomrule
    \end{tabular}
    \label{tab:lognorm-comparison}
\end{table}

Second, we looked at the performance under the lognormal partition in Tab.~\ref{tab:lognorm-comparison}. We observed that the difference between HF-MAML and FedAvg is greater than for the equal partition. While there was initially no difference in TAR@FAR0.1 for the clients added after training under the equal partition, the lognormal partition showed a larger and more significant difference between FedAvg and HF-MAML. This suggests that HF-MAML may be more effective under quantity skew. 

Another interesting result is that adding model regularization seems best when one does not tune locally afterward. Embedding regularization seems to be effective in addressing embedding drift, especially for new clients.

%\begin{figure}
%    \centering
%    \includegraphics[width=\linewidth]{sections/results/attr_results.png}
%    \caption{Results for the attributes-based partition dataset. Note that FedFR is not tuned, and thus, the bars are equivalent.}
%    \label{fig:attr_comparison}
%\end{figure}

\begin{table}
    \centering
    \caption{TAR@FAR 0.1 for the attribute-based partition dataset. Bold implies the best result.}
    \begin{tabular}{lcc}
    \toprule
    & Train & Test \\
    & \multicolumn{2}{c}{Untuned} \\
    \midrule
    FedAvg - Global & $0.718 \pm 0.007$ & $0.572 \pm 0.014$ \\
    FedAvg - Local & $0.754 \pm 0.008$ & $0.636 \pm 0.013$ \\
    FedAvg - Regularization & $0.764 \pm 0.009$ & $0.673 \pm 0.014$ \\
    HF-MAML - Global & $0.784 \pm 0.008$ & $0.653 \pm 0.015$ \\
    HF-MAML - Local & $\mathbf{0.789} \pm 0.009$ & $0.679 \pm 0.011$ \\
    HF-MAML - Regularization & $\mathbf{0.789} \pm 0.009$ & $\mathbf{0.708} \pm 0.014$ \\ \toprule
    & \multicolumn{2}{c}{Tuned} \\
    \midrule
    FedAvg - Global & $0.788 \pm 0.009$ & $0.672 \pm 0.017$ \\
    FedAvg - Local & $0.772 \pm 0.008$ & $0.695 \pm 0.013$ \\
    FedAvg - Regularization & $0.779 \pm 0.008$ & $0.694 \pm 0.015$ \\
    HF-MAML - Global & $\mathbf{0.838} \pm 0.008$ & $0.755 \pm 0.016$ \\
    HF-MAML - Local & $0.816 \pm 0.006$ & $\mathbf{0.759} \pm 0.011$ \\
    HF-MAML - Regularization & $0.812 \pm 0.006$ & $0.729 \pm 0.011$ \\ \bottomrule
    \end{tabular}
    \label{tab:attr-comparison}
\end{table}

Finally, we compared our test bench under the attribute-based partition in Tab.~\ref{tab:attr-comparison}. This partition has the same amount of data per client but has mixed attributes. This partition thus allows us to evaluate under feature skew. In this partition, we see a clear improvement from FedAvg to HF-MAML after tuning the local data. This effect is visible for both the clients known during training and the clients added after training. We also see the same effect as observed under lognormal for embedding regularization.

\subsection{Fairness Analysis}

%TODO: Verder vanaf hier!

Another vital aspect of PFR is fairness between clients. If clients know in advance that they will mostly contribute and do not receive much from collaborative training, they are unlikely to participate. To evaluate the fairness of our approaches, we note the standard deviation of our TAR@FAR 0.1 scores. The variance of the evaluation metrics is commonly used to measure the fairness of an approach.

%\begin{figure}
%    \centering
%    \includegraphics[width=0.9\linewidth]{sections/results/equal_std_results.png}
%    \caption{Standard deviation of client's global evaluation scores for the equal class partition dataset.}
%    \label{fig:equal_std_comparison}
%\end{figure}
\begin{table}
    \centering
    \caption{Standard deviation of client's evaluation scores for the equal partition dataset. Bold implies the best result.}
    \begin{tabular}{lcc}
    \toprule
    & Train & Test \\
    & \multicolumn{2}{c}{Untuned} \\
    \midrule
    FedAvg - Global & $0.031 \pm 0.005$ & $0.021 \pm 0.010$ \\
    FedAvg - Local & $0.033 \pm 0.006$ & $0.020 \pm 0.009$ \\
    HF-MAML - Global & $\mathbf{0.025} \pm 0.005$ & $0.010 \pm 0.008$ \\
    HF-MAML - Local & $0.031 \pm 0.006$ & $\mathbf{0.009} \pm 0.009$ \\ \toprule
    & \multicolumn{2}{c}{Tuned} \\
    \midrule
    FedAvg - Global & $0.027 \pm 0.005$ & $0.012 \pm 0.010$ \\
    FedAvg - Local & $0.031 \pm 0.004$ & $0.018 \pm 0.009$ \\
    HF-MAML - Global & $\mathbf{0.023} \pm 0.005$ & $\mathbf{0.007} \pm 0.008$ \\
    HF-MAML - Local & $0.028 \pm 0.005$ & $0.012 \pm 0.010$ \\ \bottomrule
    \end{tabular}
    \label{tab:equal-std-comparison}
\end{table}

In Tab.~\ref{tab:equal-std-comparison}, we noted the standard deviation of the compared methods under the equal partition. We see that local tuning of the global model improves the fairness of the model for both the FedAvg and HF-MAML models. We also see that the difference between the two under equal class partitions is small. This implies that using meta-learning when the data is approximately balanced does not add much to fairness.

%\begin{figure}
%    \centering
%    \includegraphics[width=0.9\linewidth]{sections/results/lognorm_std_results.png}
%    \caption{Standard deviation of client's global evaluation scores for the lognormal class partition dataset.}
%    \label{fig:lognorm_std_comparison}
%\end{figure}
\begin{table}
    \centering
    \caption{Standard deviation of client's evaluation scores for the lognormal partition dataset. Bold implies the best result.}
    \begin{tabular}{lcc}
    \toprule
    & Train & Test \\
    & \multicolumn{2}{c}{Untuned} \\
    \midrule
    FedAvg - Global & $\mathbf{0.121} \pm 0.022$ & $0.005 \pm 0.009$ \\
    FedAvg - Local & $0.129 \pm 0.018$ & $0.006 \pm 0.012$ \\
    FedAvg - Regularization & $0.145 \pm 0.020$ & $0.008 \pm 0.008$ \\
    HF-MAML - Global & $0.134 \pm 0.016$ & $0.006 \pm 0.008$ \\
    HF-MAML - Local & $0.124 \pm 0.021$ & $\mathbf{0.004} \pm 0.009$ \\
    HF-MAML - Regularization & $0.127 \pm 0.016$ & $0.013 \pm 0.008$ \\ \toprule
    & \multicolumn{2}{c}{Tuned} \\
    \midrule
    FedAvg - Global & $0.115 \pm 0.026$ & $0.039 \pm 0.012$ \\
    FedAvg - Local & $0.114 \pm 0.019$ & $0.029 \pm 0.011$ \\
    FedAvg - Regularization & $0.127 \pm 0.024$ & $\mathbf{0.021} \pm 0.013$ \\
    HF-MAML - Global & $\mathbf{0.088} \pm 0.028$ & $0.043 \pm 0.016$ \\
    HF-MAML - Local & $0.105 \pm 0.027$ & $0.039 \pm 0.010$ \\
    HF-MAML - Regularization & $0.107 \pm 0.022$ & $0.021 \pm 0.010$ \\ \bottomrule
    \end{tabular}
    \label{tab:lognorm-std-comparison}
\end{table}

In Tab.~\ref{tab:lognorm-std-comparison}, we show the fairness results for the lognormal partition. We see that the variance between clients is overall higher than what we observed in Tab.~\ref{tab:equal-std-comparison}, which is a logical consequence of the skew in the quantity between clients. However, this partition shows a difference between FedAvg and HF-MAML. After tuning locally, the trained clients achieve a significantly lower standard deviation than under FedAvg. Interestingly, we do not see this effect for the clients added later. 

However, we note that some of these test clients, particularly 17 and 18, possess some of the largest local datasets in our evaluation. On average, their variance is also lower than that of the clients involved in training.

%\begin{figure}
%    \centering
%    \includegraphics[width=0.9\linewidth]{sections/results/attr_std_results.png}
%    \caption{Standard deviation of client's evaluation scores for the attribute-based partition dataset.}
%    \label{fig:attr_std_comparison}
%\end{figure}
\begin{table}
    \centering
    \caption{Standard deviation of client's evaluation scores for the attribute-based partition dataset. Bold implies the best result.}
    \begin{tabular}{lcc}
    \toprule
    & Train & Test \\
    & \multicolumn{2}{c}{Untuned} \\
    \midrule
    FedAvg - Global & $0.098 \pm 0.009$ & $0.070 \pm 0.020$ \\
    FedAvg - Local & $0.099 \pm 0.008$ & $0.079 \pm 0.013$ \\
    FedAvg - Regularization & $0.103 \pm 0.010$ & $0.070 \pm 0.013$ \\
    HF-MAML - Global & $0.088 \pm 0.013$ & $0.070 \pm 0.013$ \\
    HF-MAML - Local & $\mathbf{0.083} \pm 0.009$ & $\mathbf{0.060} \pm 0.013$ \\
    HF-MAML - Regularization & $0.095 \pm 0.006$ & $0.063 \pm 0.013$ \\ \toprule
    & \multicolumn{2}{c}{Tuned} \\
    \midrule
    FedAvg - Global & $0.078 \pm 0.009$ & $0.080 \pm 0.014$ \\
    FedAvg - Local & $0.093 \pm 0.010$ & $0.068 \pm 0.013$ \\
    FedAvg - Regularization & $0.095 \pm 0.008$ & $0.072 \pm 0.008$ \\
    HF-MAML - Global & $\mathbf{0.063} \pm 0.011$ & $0.069 \pm 0.015$ \\
    HF-MAML - Local & $0.074 \pm 0.007$ & $\mathbf{0.062} \pm 0.011$ \\
    HF-MAML - Regularization & $0.081 \pm 0.008$ & $0.065 \pm 0.010$ \\ \bottomrule
    \end{tabular}
    \label{tab:attr-std-comparison}
\end{table}

\begin{figure}[t]
    \centering
\begin{tikzpicture}
    \begin{axis}[
        xbar=0pt,
        bar width=6pt,
        height=13cm,
        width=0.45\textwidth,
        enlarge y limits=0.05,
        xlabel={TAR@FAR0.1},
        ylabel={Client ID},
        symbolic y coords={1,2,3,4,5,6,7,8,9,10,11,12,13,14,15,16,17,18,19,20},
        ytick=data,
        y tick label style={rotate=0, anchor=east},
        legend style={at={(0.5,-0.08)},
        anchor=north,legend columns=2},
        xmin=0,
        xmax=1.1
    ]

    \addplot[fill=red!70] coordinates {(0.81094,1) (0.8561,2) (0.88282,3) (0.8509,4) 
                          (0.773,5) (0.72204,6) (0.89308,7) (0.8748,8) 
                          (0.82798,9) (0.90928,10) (0.93414,11) (0.90006,12) 
                          (0.78584,13) (0.82738,14) (0.72056,15) (0.85068,16) 
                          (0.73512,17) (0.7651,18) (0.78424,19) (0.64044,20)};
  \addplot[fill=gray!40] coordinates {(0.76186,1) (0.81474,2) (0.83876,3) (0.82918,4) 
                          (0.71984,5) (0.66762,6) (0.84204,7) (0.82076,8) 
                          (0.7731,9) (0.875,10) (0.89504,11) (0.8733,12) 
                          (0.6994,13) (0.78506,14) (0.61772,15) (0.79126,16) 
                          (0.64256,17) (0.66112,18) (0.71394,19) (0.5506,20)};
    
    \addplot[black, only marks, nodes near coords, 
             point meta=explicit symbolic, every node near coord/.append style={anchor=west, shift={(0pt,0pt)}}] 
    %coordinates {(0.81094,1) [+6.4\%] (0.8561,2) [+5.1\%] (0.88282,3) [+5.3\%] 
    %             (0.8509,4) [+2.6\%] (0.773,5) [+7.4\%] (0.72204,6) [+8.2\%] 
    %             (0.89308,7) [+6.1\%] (0.8748,8) [+6.6\%] (0.82798,9) [+7.1\%] 
    %             (0.90928,10) [+3.9\%] (0.93414,11) [+4.4\%] (0.90006,12) [+3.1\%] 
    %             (0.78584,13) [+12.4\%] (0.82738,14) [+5.4\%] (0.72056,15) [+16.6\%] 
    %             (0.85068,16) [+7.5\%] (0.73512,17) [+14.4\%] (0.7651,18) [+15.7\%] 
     %            (0.78424,19) [+9.8\%] (0.64044,20) [+16.3\%]};
         coordinates {(0.9,1) [+6.4\%] (0.9,2) [+5.1\%] (0.9,3) [+5.3\%] 
                 (0.9,4) [+2.6\%] (0.9,5) [+7.4\%] (0.9,6) [+8.2\%] 
                 (0.9,7) [+6.1\%] (0.9,8) [+6.6\%] (0.9,9) [+7.1\%] 
                 (0.9,10) [+3.9\%] (0.9,11) [+4.4\%] (0.9,12) [+3.1\%] 
                 (0.9,13) [+12.4\%] (0.9,14) [+5.4\%] (0.9,15) [+16.6\%] 
                 (0.9,16) [+7.5\%] (0.9,17) [+14.4\%] (0.9,18) [+15.7\%] 
                 (0.9,19) [+9.8\%] (0.9,20) [+16.3\%]};
                 
    \legend{HF-MAML,FedAvg}
    \end{axis}
\end{tikzpicture}
    \caption{The mean TAR@FAR0.1 per client for FedAvg and HFMAML using the local approach on the attributes partition. Percentage improvement by HFMAML is presented at the end of the bar. Clients with weak performance on average have a greater improvement with HFMAML.}
    \label{fig:attr-var-client}
\end{figure}
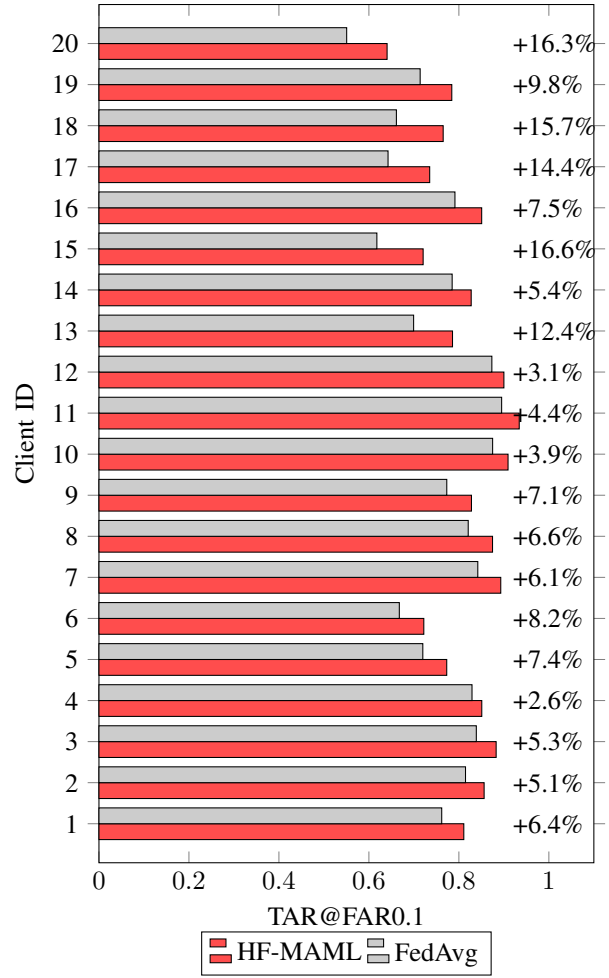

Finally, we noted the fairness results for the attribute-based partition in Tab.~\ref{tab:attr-std-comparison}. Here, we observe the largest difference between FedAvg and HF-MAML. The difference is small but significant for the 15 trained clients before tuning. After tuning, the standard deviation of HF-MAML becomes smaller, and the difference between FedAvg and HF-MAML becomes even larger. Under feature skew, HF-MAML seems to achieve the fairest results.

Finally, to better evaluate why our HF-MAML-based approach achieved fairer results, we compared the per-client performance for both FedAvg and HF-MAML in Fig.~\ref{fig:attr-var-client}. We observed that HF-MAML achieved a higher TAR@FAR 0.1 for all clients. However, the weakest clients under FedAvg achieved the greatest improvement percentage-wise with HF-MAML. This indicates that HF-MAML gives fairer results primarily by improving the performance of the weakest clients. Note that the last $5$ clients were not seen during training, which implies that HF-MAML is a better alternative when new clients also want a personalized model.

\section{Discussion \& Conclusion}
In general, our results indicate that the effectiveness of HF-MAML varies depending on the level and type of data heterogeneity. Under the equal partition, HF-MAML was similar to FedAvg. Thus, it is hard to justify the additional cost and complexity of approximating the second derivative in this scenario. However, if we assume that there is some form of data heterogeneity, HF-MAML is beneficial compared to FedAvg. We observe that, especially after tuning locally, HF-MAML achieves better local TAR@FAR scores and lower variance between clients compared to FedAvg. 

When it comes to embedding regularization, we observe that it helps to transfer the model to new clients. By not allowing the model to drift too much, we avoid a model that is too specific to a set of clients and, thus, transfers better to new clients. This improvement is not retained under local tuning, so regularization is primarily interesting when local tuning is not an option.

In conclusion, we propose using HF-MAML in the context of FFR. We proposed new data partitions based on the CelebA dataset that help evaluate FR in a data-heterogeneous setting and demonstrated that HF-MAML can add value in such settings. The limitations of the work entail the limited performance of our approach. By not including a global dataset, our work is unable to reach practically useful TAR@FAR scores. Our work is intended as an investigation of whether HF-MAML can be useful in FFR and should provide a stepping stone in cases where a global dataset is no longer feasible. Future work should also look at applying our approach to more datasets with different types of data heterogeneity.
\printbibliography

\end{document}